\documentclass{article}

\usepackage{PRIMEarxiv}

\usepackage{multirow}

\usepackage[utf8]{inputenc} 
\usepackage[T1]{fontenc}    
\usepackage{hyperref}       
\usepackage{url}            
\usepackage{booktabs}       
\usepackage{amsfonts}       
\usepackage{nicefrac}       
\usepackage{microtype}      
\usepackage{lipsum}
\usepackage{fancyhdr}       
\usepackage{graphicx}       
\graphicspath{{media/}}     

\title{DB SwinT: A Dual-Branch Swin Transformer Network for Road Extraction in Optical Remote Sensing Imagery
}

\author{
\textbf{Zongyang He} \\
School of Information Science and Engineering \\
Chongqing Jiaotong University \\
Chongqing 400074, China \\
hezongyang@mails.cqjtu.edu.cn
\and
\textbf{Xiangli Yang} \\
School of Information Science and Engineering \\
Chongqing Jiaotong University \\
Chongqing 400074, China \\
xiangli.yang@cqjtu.edu.cn
\and
\\
\textbf{Xian Gao} \\
College of Information Engineering \\
Inner Mongolia University of Technology \\
Inner Mongolia Key Laboratory\\ 
of Radar Technology and Application \\
Hohhot 010051, China \\
gaoxian\_gx@163.com
\and
\\
\textbf{Zhiguo Wang} \\
College of Information Engineering \\
Inner Mongolia University of Technology \\
Inner Mongolia Key Laboratory\\ 
of Radar Technology and Application \\
Hohhot 010051, China \\
wangzhiguo@imut.edu.cn
}

\begin{document}
\maketitle

\begin{abstract}
With the continuous improvement in the spatial resolution of optical remote sensing imagery, accurate road extraction has become increasingly important for applications such as urban planning, traffic monitoring, and disaster management. However, road extraction in complex urban and rural environments remains challenging, as roads are often occluded by trees, buildings, and other objects, leading to fragmented structures and reduced extraction accuracy. To address this problem, this paper proposes a Dual-Branch Swin Transformer network (DB SwinT) for road extraction. The proposed framework combines the long-range dependency modeling capability of the Swin Transformer with the multi-scale feature fusion strategy of U-Net, and employs a dual-branch encoder to learn complementary local and global representations. Specifically, the local branch focuses on recovering fine structural details in occluded areas, while the global branch captures broader semantic context to preserve the overall continuity of road networks. In addition, an Attentional Feature Fusion (AFF) module is introduced to adaptively fuse features from the two branches, further enhancing the representation of occluded road segments. Experimental results on the Massachusetts and DeepGlobe datasets show that DB SwinT achieves Intersection over Union (IoU) scores of 79.35\% and 74.84\%, respectively, demonstrating its effectiveness for road extraction from optical remote sensing imagery.
\end{abstract}


\section{Introduction}
The rapid development of optical remote sensing technology has expanded the application of road extraction in areas such as urban planning \cite{ref1}, geographic information system (GIS) updating \cite{ref2}, and autonomous driving \cite{ref3}. Traditional road data acquisition methods mainly rely on ground surveys, which are inefficient for large-scale and timely updates. Therefore, automatic road extraction from optical remote sensing imagery has become an important research topic in remote sensing \cite{ref4,ref5}. In complex urban environments, dense buildings and vegetation often occlude roads, resulting in fragmented or discontinuous structures in the imagery. In rural regions, roads are also frequently covered by natural objects such as trees, which further reduces extraction accuracy. Consequently, achieving reliable road extraction under complex backgrounds and occlusion conditions remains a major research challenge \cite{ref6,ref7}.

In recent years, rapid progress in deep learning, especially convolutional neural networks (CNNs), has significantly improved road extraction performance. To overcome the limited ability of early convolutional models in capturing multi-scale contextual information, U-Net \cite{ref8} and its variants adopted an encoder–decoder architecture with skip connections to integrate features at different resolutions. UNet++ \cite{ref9} further introduced dense skip pathways to reduce the semantic gap between encoder and decoder features and to enhance multi-scale representation. To alleviate performance degradation in deeper networks and facilitate feature propagation, Res-UNet \cite{ref10} incorporated residual connections, which help stabilize gradient flow and improve feature learning. To address background interference and occlusion in complex scenes, Attention U-Net \cite{ref11} introduced attention mechanisms that highlight informative regions and suppress irrelevant features. To strengthen the representation of overall road structures through the stacking of multiple convolutional layers, Tan et al. \cite{ref12} constructed an end-to-end convolutional network. To alleviate the issue of limited labeled samples, Wei et al. \cite{ref13} introduced a domain adaptation-based image synthesis strategy. To achieve a balance between computational efficiency and segmentation accuracy, LinkNet \cite{ref14} employed a lightweight encoder–decoder framework with residual connections, reducing model complexity while maintaining performance. In addition, GAMSNet \cite{ref15} combined multi-scale feature extraction with attention mechanisms to improve robustness to scale and texture variations. To deal with road discontinuities caused by occlusions and complicated backgrounds, Wan et al. \cite{ref16} proposed DA-RoadNet, which models spatial and channel dependencies using a dual-attention strategy. Luo et al. \cite{ref17} further presented AD-RoadNet, introducing hybrid receptive field and topological reasoning modules to improve road connectivity in occluded scenes.

Although CNN-based methods have achieved strong performance, their inherently local receptive fields restrict their ability to capture global context and long-range dependencies, especially in large-scale scenes with severe occlusions \cite{ref18}. Increasing network depth or enlarging receptive fields can partly mitigate this limitation, but such approaches usually introduce higher computational costs and may increase the risk of overfitting. To address these issues, Transformer architectures have recently been introduced into computer vision tasks \cite{ref19}. Initially developed for natural language processing \cite{ref20}, Transformers employ self-attention mechanisms that are well suited for modeling long-range relationships. Vision Transformer (ViT) \cite{ref21} demonstrated that Transformer architectures can be effectively applied to image classification, while SETR \cite{ref22} further revealed the potential of pure Transformer models for image segmentation. To improve computational efficiency, Swin Transformer \cite{ref23} proposed a window-based self-attention mechanism, achieving state-of-the-art results across a variety of vision tasks. For road extraction applications, BDTNet \cite{ref24} integrates a bidirectional Transformer structure to jointly model global context and local details. Yang et al. \cite{ref25} adopted Transformer-based frameworks to enhance robustness under occlusion conditions, while Light Roadformer \cite{ref26} focuses on efficient road extraction with reduced computational overhead. In addition, RoadFormer \cite{ref27}, built on the Swin Transformer backbone, improves road extraction accuracy under complex scenarios such as varying road widths and occlusions.

Despite these advances, effectively combining multi-scale feature representation with global contextual modeling remains a challenging issue for Transformer-based road extraction methods. To overcome this limitation, this paper presents a Dual-Branch Swin Transformer (DB SwinT) network designed for road extraction from optical remote sensing imagery. The primary contributions of this study are summarized as follows:

\begin{itemize}
\item A novel Dual-Branch Swin Transformer framework is proposed in this work, which combines the long-range dependency modeling capability of the Swin Transformer with the multi-scale feature fusion strength of U-Net.
\item A dual-branch U-shaped encoder is developed, where the local branch recovers fine-grained road structures in occluded regions, and the global branch captures semantic context to preserve road connectivity.
\item An Attentional Feature Fusion (AFF) module is introduced to adaptively fuse local and global features, improving their complementarity and enhancing road extraction performance in scenes with complex occlusions.
\end{itemize}

\section{Methodology}\label{sec:methodology}
\begin{figure}
\centering
\includegraphics[scale=.5]{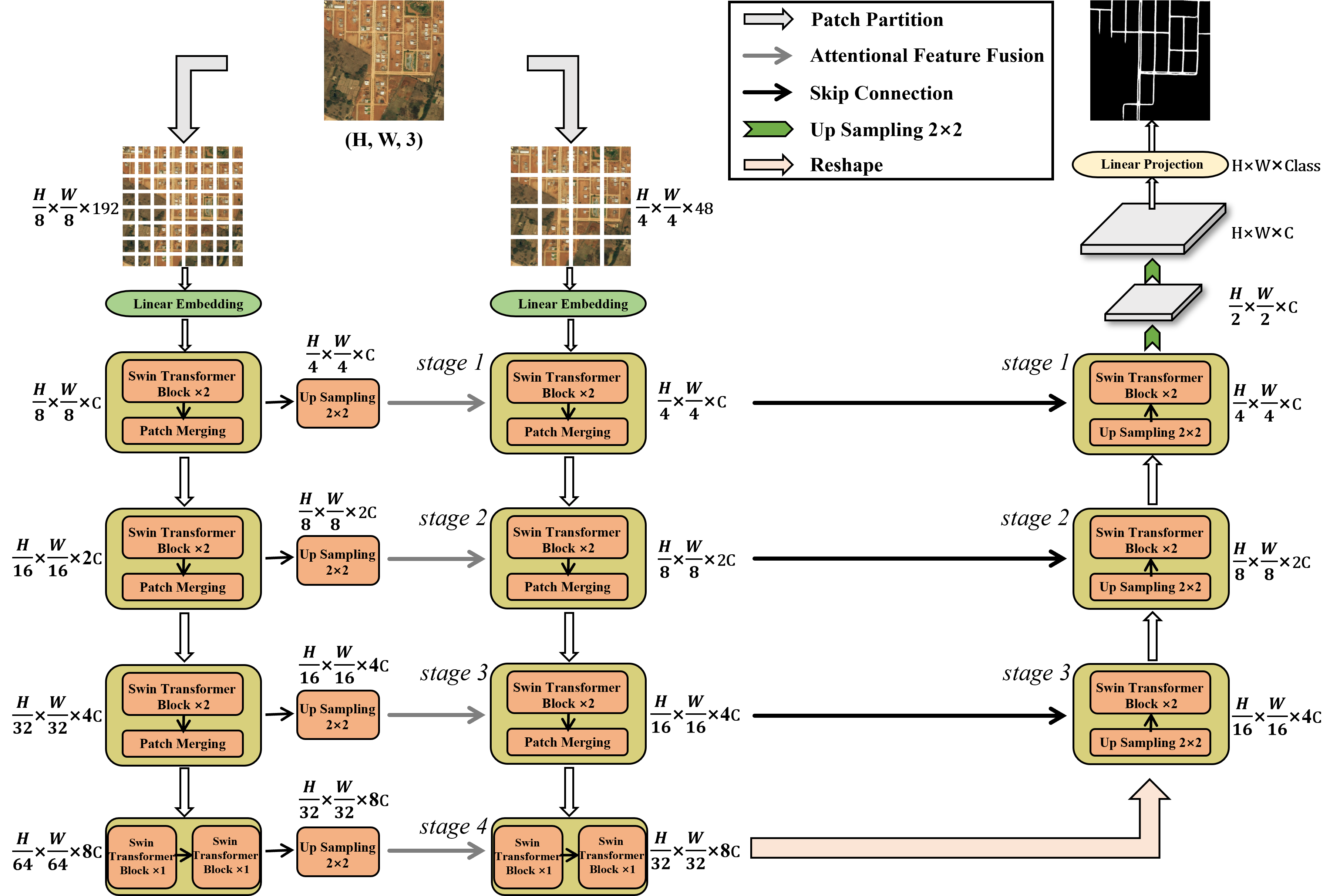}
\caption{Overall architecture of DB SwinT, including dual-branch encoder, decoder, AFF module, and skip connections.}
\label{fig1}
\end{figure}

This study addresses road extraction by combining the long-range dependency modeling capability of the Swin Transformer with the multi-scale feature fusion strategy of U-Net. A dual-branch encoder is designed to learn both local and global representations. As illustrated in Figure \ref{fig1}, the framework consists of a dual-branch encoder, a decoder, an AFF module, and skip connections. The input optical remote sensing image is first divided into non-overlapping patches at different scales and fed into two encoder branches to capture complementary local and global features. These features are then aligned to the same spatial resolution and fused through the AFF module. The decoder subsequently upsamples the fused features using Swin Transformer blocks and skip connections to incorporate multi-scale information from the encoder. Finally, the feature maps are restored to the original image resolution to produce the road extraction results, and the detailed structure of each component is described in the following subsections.

\subsection{Swin Transformer Block}\label{subsec:swin}
\begin{figure}
\centering
\includegraphics[scale=.55]{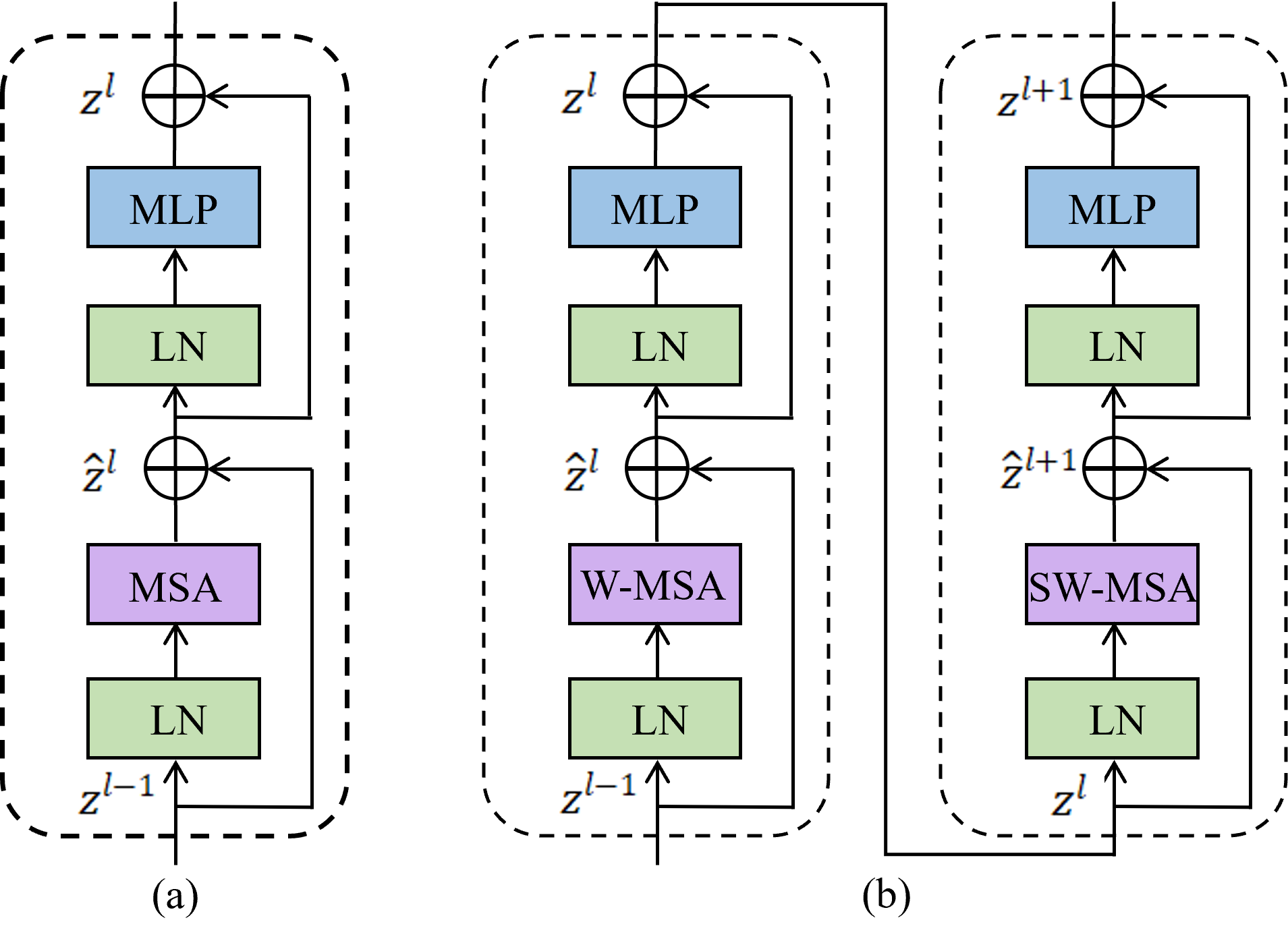}
\caption{Illustration of (a) standard Transformer and (b) Swin Transformer blocks.}
\label{fig2}
\end{figure}

The standard Transformer encoder~\cite{ref20} is composed of $L$ stacked layers, where each layer includes a Multi-Head Self-Attention (MSA) module and a Multi-Layer Perceptron (MLP). Layer Normalization (LN) is applied before each module, followed by residual connections to stabilize training, as shown in Figure \ref{fig2}(a). The output of the $l$-th layer can be formulated as:

\begin{equation}
\hat{z}^{(l)} = \mathrm{MSA}(\mathrm{LN}(z^{(l-1)})) + z^{(l-1)},
\end{equation}
\begin{equation}
z^{(l)} = \mathrm{MLP}(\mathrm{LN}(\hat{z}^{(l)})) + \hat{z}^{(l)}.
\end{equation}

In the standard Transformer architecture, each token interacts with all other tokens, leading to quadratic computational complexity, which is inefficient for dense prediction tasks. To address this issue, Swin Transformer~\cite{ref23} introduces window-based Multi-Head Self-Attention (W-MSA) and Shifted Window-based Self-Attention (SW-MSA). Specifically, W-MSA partitions the input features into non-overlapping local windows of $M \times M$ patches (with a default size of 7) and performs self-attention only within each window, thereby reducing computational cost. This process is illustrated in Figure \ref{fig2}(b), and the computation can be expressed as:

\begin{equation}
\hat{z}^{(l)} = \mathrm{W-MSA}(\mathrm{LN}(z^{(l-1)})) + z^{(l-1)},
\end{equation}
\begin{equation}
z^{(l)} = \mathrm{MLP}(\mathrm{LN}(\hat{z}^{(l)})) + \hat{z}^{(l)}.
\end{equation}

The main limitation of W-MSA is the lack of cross-window information exchange. To address this without increasing computational cost, SW-MSA cyclically shifts the feature map top-left, enabling efficient cross-window interaction. After the shift, each batch window contains multiple non-adjacent sub-windows, while the total number of windows remains unchanged. In both W-MSA and SW-MSA, self-attention within local windows incorporates relative position bias. Based on this design, the outputs of SW-MSA and the MLP module are formulated as:

\begin{equation}
\hat{z}^{(l+1)} = \mathrm{SW-MSA}(\mathrm{LN}(z^{(l)})) + z^{(l)},
\end{equation}
\begin{equation}
z^{(l+1)} = \mathrm{MLP}(\mathrm{LN}(\hat{z}^{(l+1)})) + \hat{z}^{(l+1)}.
\end{equation}

\subsection{Encoder}\label{subsec:encoder}
The overall architecture follows the U-shaped structure of U-Net~\cite{ref8}, with the encoder built using Swin Transformer blocks. As illustrated in Figure \ref{fig1}, the input image of size $H \times W$ is divided into non-overlapping patches of size $S \times S$ and linearly projected into a $C$-dimensional embedding space. The resulting embeddings are then processed through four encoder stages, each consisting of two Swin Transformer blocks.

To capture hierarchical features, patch merging layers are introduced in the first three stages to gradually reduce spatial resolution while increasing channel dimensions. In each patch merging step, features from neighboring $2 \times 2$ patches are aggregated and followed by a linear projection, producing a $2\times$ down-sampling and doubling the number of channels. As a result, the feature resolutions across the four stages become $\frac{H}{S}\times\frac{W}{S}$, $\frac{H}{2S}\times\frac{W}{2S}$, $\frac{H}{4S}\times\frac{W}{4S}$, and $\frac{H}{8S}\times\frac{W}{8S}$, with channel dimensions of $C$, $2C$, $4C$, and $8C$, respectively. The final stage keeps the same resolution to prevent convergence problems that may arise from excessive down-sampling.

\subsection{Decoder}\label{subsec:decoder}
As shown in Figure \ref{fig1}, the decoder contains three stages, each including up-sampling, skip connections, and Swin Transformer blocks. The feature map from the fourth encoder stage serves as the initial input to the decoder. In each stage, features are first up-sampled by a factor of 2 and concatenated with the corresponding encoder feature maps through skip connections, followed by a Swin Transformer block for feature refinement. This design allows the decoder to integrate multi-scale information from the encoder while capturing long-range dependencies and global context, thereby improving decoding performance. After the three stages, feature maps with a resolution of $\frac{H}{4} \times \frac{W}{4}$ are obtained. Directly applying a 4$\times$ up-sampling may result in the loss of shallow features. Therefore, two successive up-sampling operations are employed to recover low-level features with resolutions of $\frac{H}{2} \times \frac{W}{2}$ and $H \times W$, ensuring that the final output matches the original image resolution. The predicted result is finally generated through a linear projection layer.

\subsection{Dual-Branch Representation}\label{subsec:dual}
The self-attention mechanism captures long-range relationships among patches. However, patch partitioning ignores fine-grained pixel details, which may lead to the loss of shallow feature information. Vision Transformer (ViT)~\cite{ref21} can improve performance in some cases by using smaller patch sizes, yet a single-scale patching strategy limits feature extraction. To enhance segmentation performance and robustness, a multi-scale Swin Transformer is employed for feature extraction. Patches of different scales provide complementary information, where smaller patches capture fine details and larger patches capture global features. Inspired by CrossViT~\cite{ref28} and DS-TransUNet~\cite{ref29}, a dual-branch Swin Transformer encoder is proposed. Specifically, the encoder consists of two independent branches: (i) the local feature extraction branch, with patch size $s=4$, focuses on occluded regions and outputs features at resolutions of $\frac{H}{4} \times \frac{W}{4}$, $\frac{H}{8} \times \frac{W}{8}$, $\frac{H}{16} \times \frac{W}{16}$, and $\frac{H}{32} \times \frac{W}{32}$; (ii) the global feature extraction branch, with patch size $s=8$, captures global semantic information and outputs features at resolutions of $\frac{H}{8} \times \frac{W}{8}$, $\frac{H}{16} \times \frac{W}{16}$, $\frac{H}{32} \times \frac{W}{32}$, and $\frac{H}{64} \times \frac{W}{64}$. This dual-branch design enables the model to extract features at multiple spatial levels, improving segmentation performance and adaptability in complex scenes.

\subsection{Attentional Feature Fusion}\label{subsec:aff}
\begin{figure}
\centering
\includegraphics[scale=1]{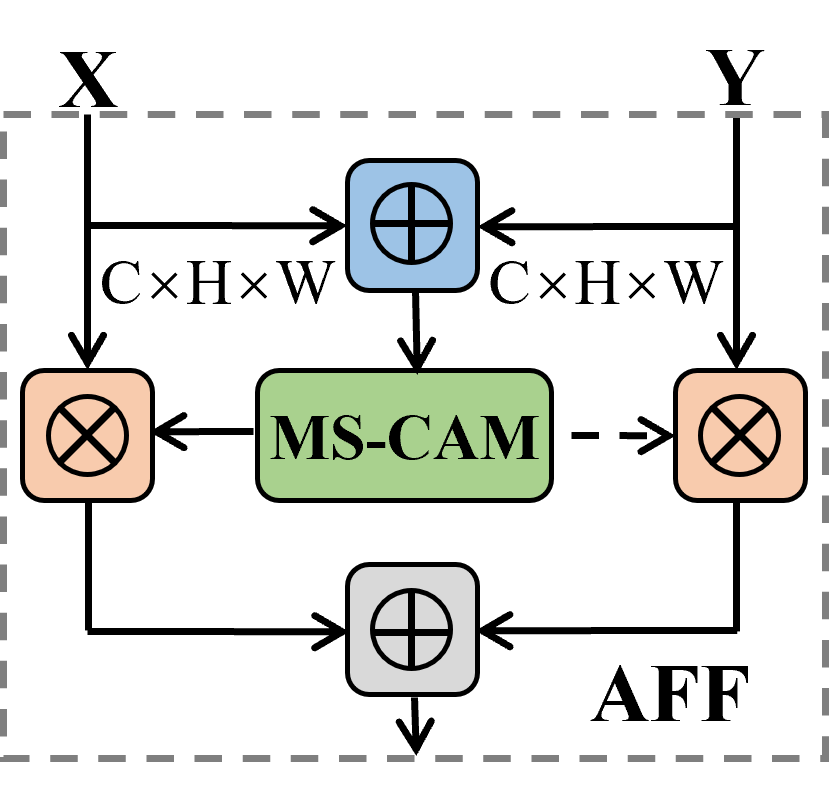}
\caption{Attentional Feature Fusion with multi-scale channel attention mechanism.}
\label{fig3}
\end{figure}

After obtaining features from the dual-branch encoder, an important challenge is effectively fusing them to maximize information utilization from both branches. Directly concatenating multi-scale feature maps followed by convolution is limited in capturing long-range dependencies, global context, and cross-scale interactions, as such operations mainly model local correlations and treat all channels uniformly without adaptive weighting. To address this issue, an Attentional Feature Fusion (AFF) module~\cite{ref30} is adopted, which employs a multi-scale channel attention mechanism to adaptively fuse features from different receptive fields and hierarchical levels. The module enhances discriminative feature responses by explicitly modeling channel-wise importance across scales, enabling better preservation of fine-grained and high-level semantic information. Specifically, the AFF module first performs an initial fusion of input features $X$ and $Y$, then extracts global contextual descriptors through channel aggregation, computes attention weights, and generates the fused feature $Z$ via a soft selection mechanism, ensuring both detailed local structures and global semantic consistency are effectively retained. The formula is as follows:

\begin{equation}
Z = M(X \oplus Y) \otimes X + \bigl(1 - M(X \oplus Y)\bigr) \otimes Y,
\end{equation}

where \( Z \in \mathbb{R}^{C \times H \times W} \) and \( \oplus \) denotes the initial feature fusion. As shown in Figure \ref{fig3}, the dashed line corresponds to \( 1 - M(X \oplus Y) \). By dynamically adjusting fusion weights \( M(X \oplus Y) \) and \( 1 - M(X \oplus Y) \), the network adaptively balances contributions from $X$ and $Y$. In practice, global branch features ($s=8$) are up-sampled to match the local branch ($s=4$) resolution, enabling effective fusion at each spatial level through skip connections. This ensures full utilization of multi-scale features from the encoder, improving model performance in complex scenes.

\subsection{Skip Connections}\label{subsec:skip}
The model follows a U-Net style architecture with skip connections to combine multi-scale encoder features with the up-sampled features from the decoder, thereby reducing spatial information loss caused by down-sampling. In particular, skip connections create direct links between encoder and decoder layers with the same spatial resolution. During the decoding process, the up-sampled feature maps are concatenated with encoder features at the corresponding scale, enabling the network to merge detailed spatial information with high-level semantic representations. This design alleviates the loss of boundary and texture information caused by successive pooling or patch merging operations in the encoder.In addition to preserving spatial resolution, skip connections facilitate gradient propagation by providing shorter information paths between shallow and deep layers, thereby stabilizing network training. The combination of low-level features containing edge and contour information with semantically rich deep features further improves the model’s capability to delineate object boundaries and maintain structural continuity. Consequently, the network achieves improved segmentation accuracy, better edge preservation, stronger global contextual awareness, and enhanced robustness when handling complex or fine-grained structures.

\section{Experiments}
\subsection{Dataset Description}
Two public datasets are used for evaluation: the DeepGlobe dataset~\cite{ref31} and the Massachusetts Roads dataset~\cite{ref32}, both consisting of optical remote sensing images covering urban, rural, and tropical rainforest environments.

\begin{figure}
\centering
\includegraphics[scale=.86]{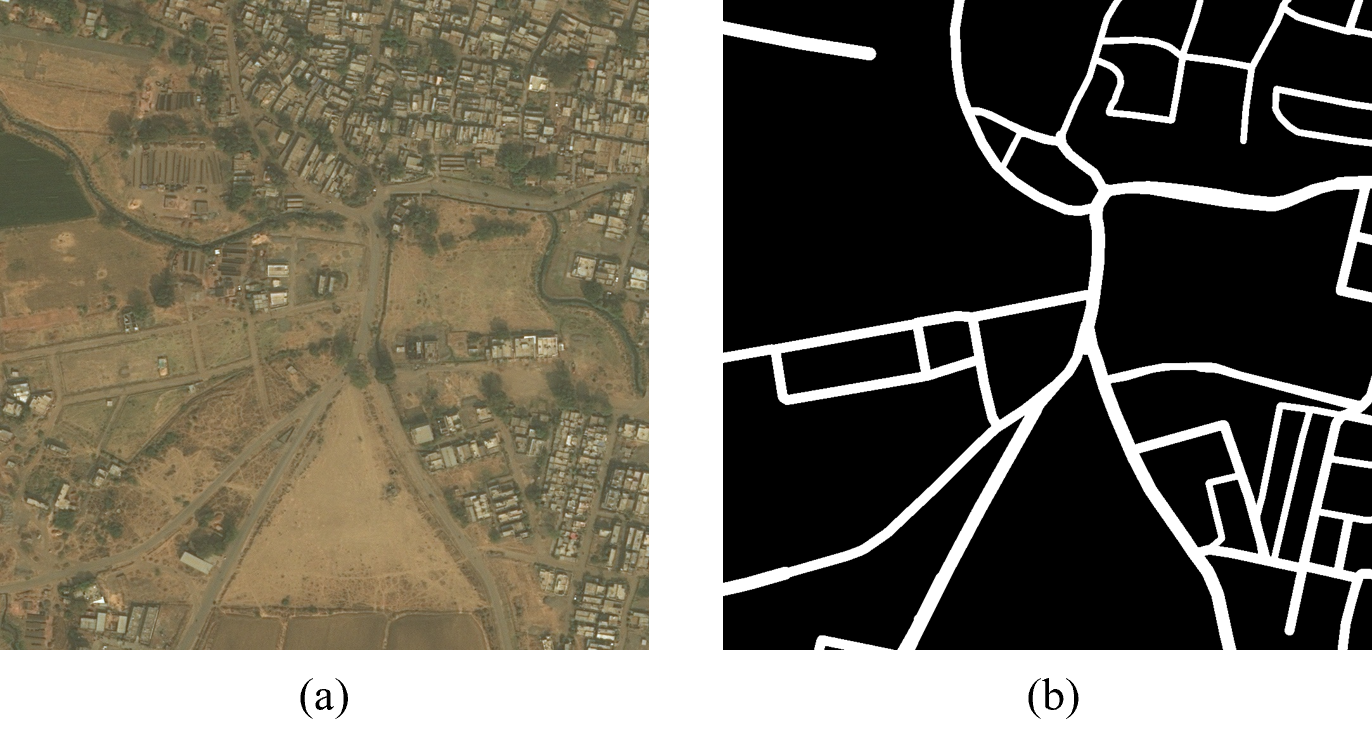}
\caption{DeepGlobe dataset. (a) Image; (b) Ground truth.}
\label{fig4}
\end{figure}

The DeepGlobe dataset, introduced as part of the DigitalGlobe satellite imagery challenge, offers high-resolution remote sensing images with a spatial resolution of 0.5 m/pixel. It comprises 6,226 image–label pairs of size $1024 \times 1024$ pixels, covering roughly 1,634 km$^2$. Road annotations are provided as binary masks for pixel-level supervision. The dataset encompasses a variety of environments, including urban, rural, and tropical rainforest areas from countries such as India, Thailand, and Indonesia. As a result, road appearances differ considerably in terms of width, surface material, lighting conditions, and surrounding land cover. In rural regions, roads are frequently occluded by vegetation or exhibit low contrast against the background, which complicates accurate boundary delineation and connectivity preservation. The dataset is randomly split into training, validation, and test sets at a ratio of 8:1:1. A sample image and its corresponding ground truth are shown in Figure \ref{fig4}.

\begin{figure}
\centering
\includegraphics[scale=.86]{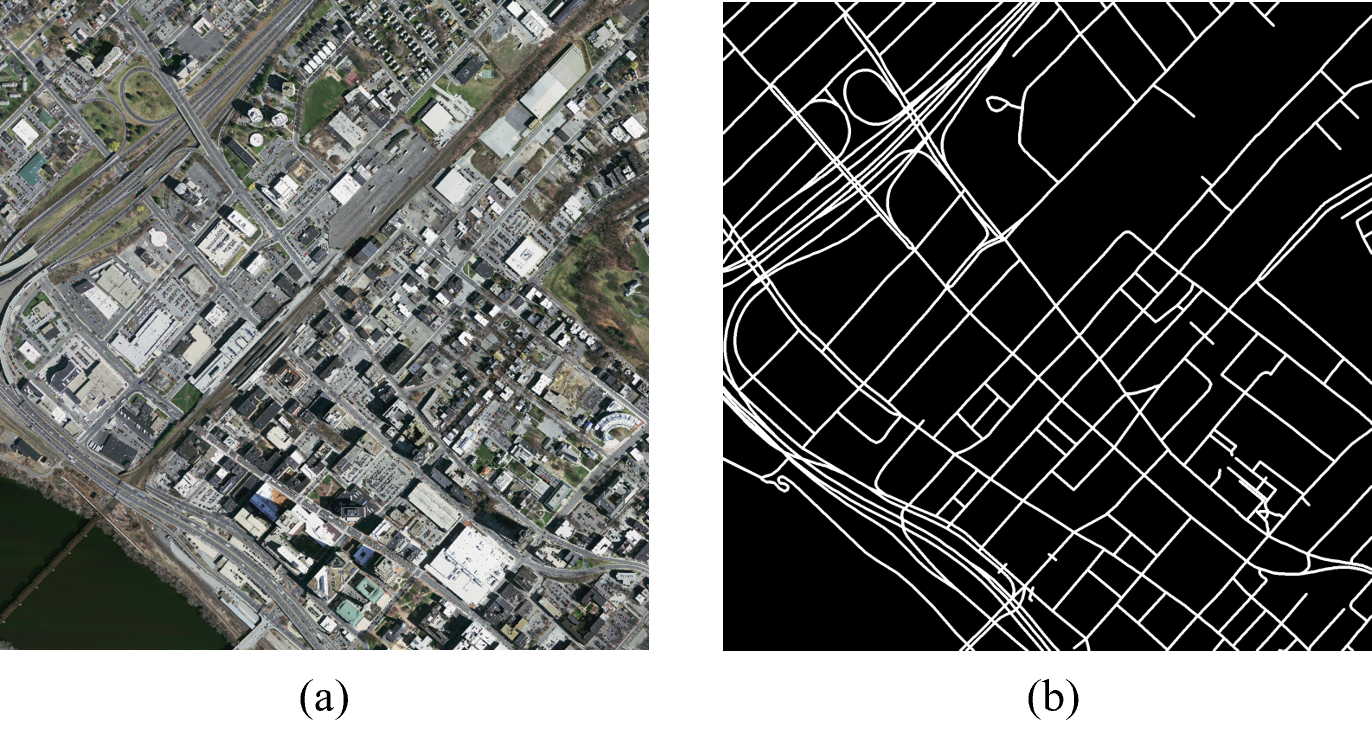}
\caption{Massachusetts dataset. (a) Image; (b) Ground truth.}
\label{fig5}
\end{figure}

The Massachusetts Roads dataset consists of 1,171 aerial image–label pairs with a spatial resolution of 1 m/pixel and image dimensions of $1500 \times 1500$ pixels, covering over 2,600 km$^2$ in Massachusetts, United States. Road annotations are provided as pixel-level binary masks for supervised segmentation. The dataset includes a variety of urban, suburban, and rural scenarios, featuring roads of different scales and densities. Compared with DeepGlobe, it contains denser built-up areas and more regular road networks, where roads are often occluded by buildings, shadows, or vehicles. Narrow streets and complex intersections further increase structural ambiguity, complicating fine-grained extraction and the maintenance of road connectivity. The dataset is divided into training, validation, and test sets in an 8:1:1 ratio to ensure consistent evaluation. Representative examples are shown in Figure \ref{fig5}.

\subsection{Network Configuration}
All experiments are carried out on a workstation running Windows 10, equipped with an Intel Core i5-12400F CPU and an NVIDIA GeForce RTX 3060 GPU. The proposed network is implemented in the PyTorch 1.11 framework. Network parameters are optimized using stochastic gradient descent (SGD) with a momentum of 0.9. The initial learning rate is set to $2 \times 10^{-4}$, with a weight decay of $2 \times 10^{-4}$. The batch size is fixed at 4, and the network is trained for 100 epochs. A step-wise learning rate decay is applied, reducing the learning rate to 20\% of its initial value every 20 epochs, which helps ensure stable convergence during training.

\subsection{Evaluation Metrics}
Road extraction is quantitatively assessed using Precision, Recall, F1-score, and Intersection over Union (IoU), all computed from pixel-level true positives, false positives, and false negatives. Precision indicates the proportion of correctly predicted road pixels among all predicted road pixels, while Recall represents the proportion of correctly identified road pixels among all ground-truth road pixels. The F1-score, calculated as the harmonic mean of Precision and Recall, provides a balanced measure of detection accuracy. IoU evaluates the overlap between the predicted road regions and ground-truth masks, reflecting overall segmentation quality. The metrics are defined as follows:

\begin{equation}
\mathrm{Precision} = \frac{TP}{TP + FP},
\end{equation}

\begin{equation}
\mathrm{Recall} = \frac{TP}{TP + FN},
\end{equation}

\begin{equation}
\mathrm{F1{-}score} = \frac{2 \times \mathrm{Precision} \times \mathrm{Recall}}
{\mathrm{Precision} + \mathrm{Recall}},
\end{equation}

\begin{equation}
\mathrm{IoU} = \frac{TP}{TP + FP + FN},
\end{equation}

\subsection{Comparative Experiments}
\begin{figure}
\centering
\includegraphics[scale=.55]{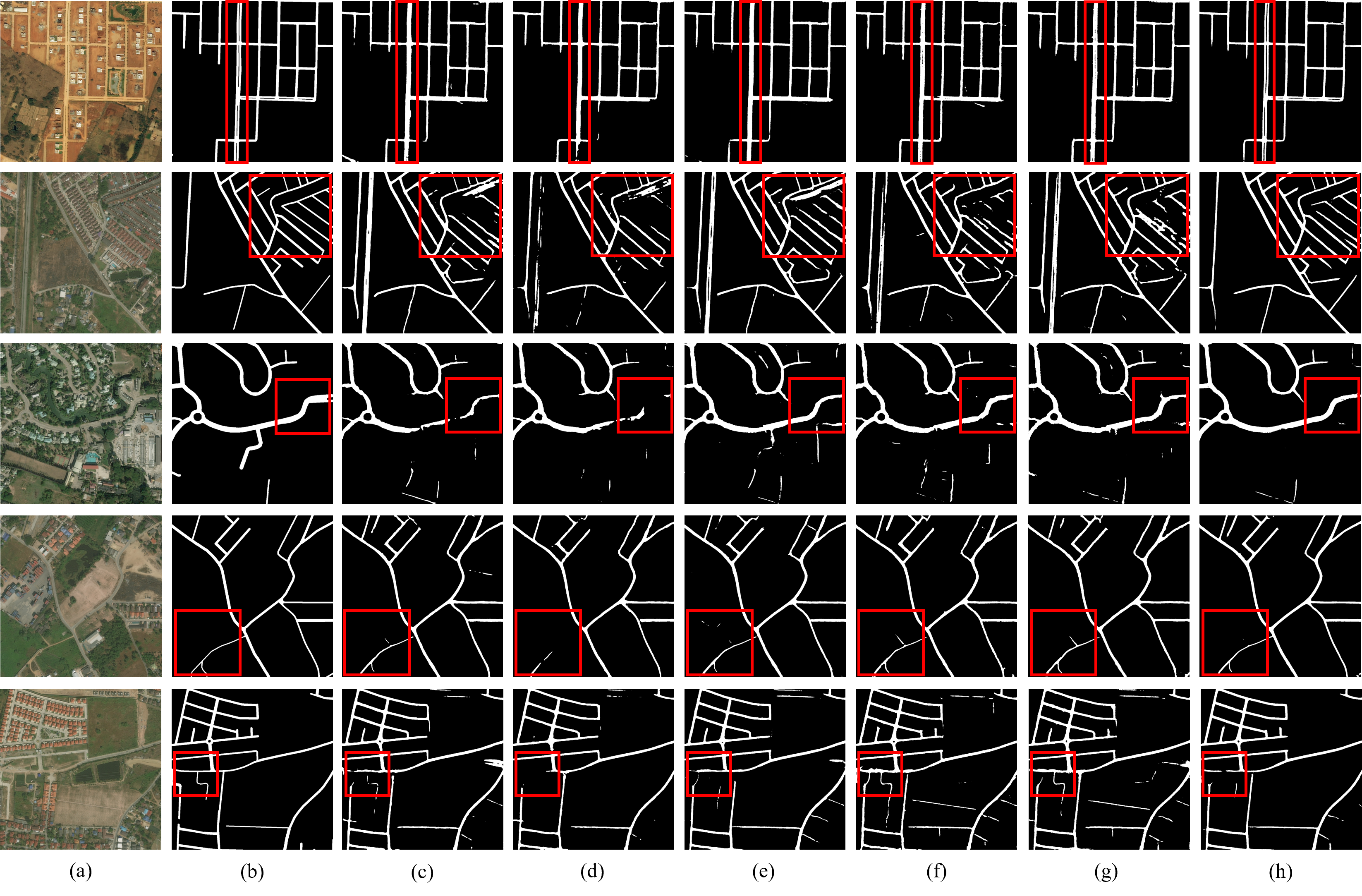}
\caption{Qualitative comparison of road extraction results on the DeepGlobe dataset.
(a) image; (b) Ground truth; (c) LinkNet50; (d) U-Net; (e) GAMSNet;
(f) SwinT; (g) SwinT+U-Net; (h) Proposed.}
\label{fig6}
\end{figure}

\begin{figure}
\centering
\includegraphics[scale=.55]{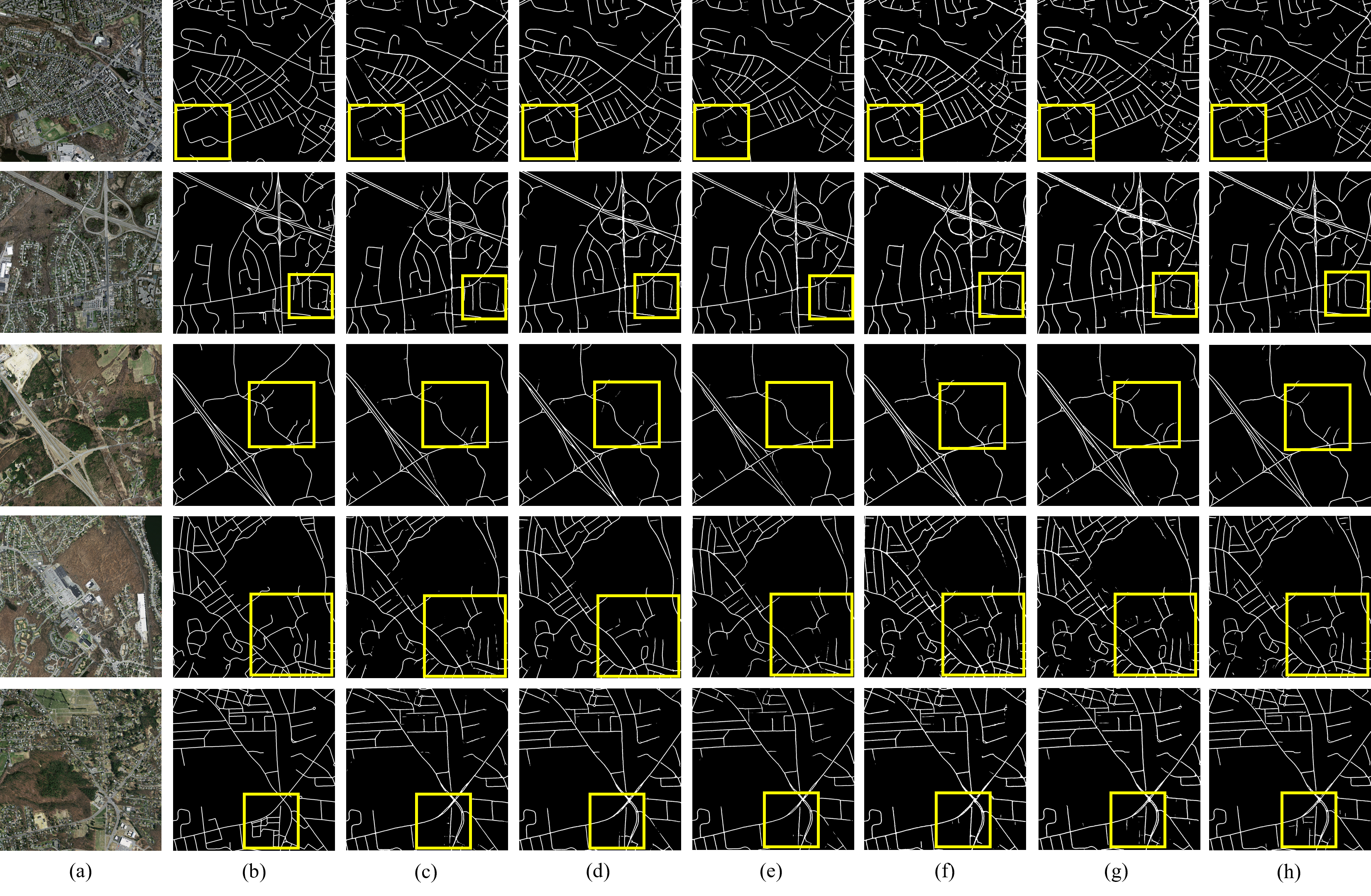}
\caption{Qualitative comparison of road extraction results on the Massachusetts dataset.
(a)image; (b) Ground truth; (c) LinkNet50; (d) U-Net; (e) GAMSNet;
(f) SwinT; (g) SwinT+U-Net; (h) Proposed.}
\label{fig7}
\end{figure}

To assess the effectiveness of the proposed DB SwinT for road extraction, experiments were conducted on DeepGlobe~\cite{ref31} and Massachusetts~\cite{ref32} datasets, representing rural and urban environments. The method was compared with four state-of-the-art models using Precision, Recall, F1-score, and IoU metrics, providing a comprehensive quantitative assessment of detection accuracy, structural completeness, and overall segmentation performance. Quantitative results are summarized in Table \ref{tab1} and Table \ref{tab2}, while qualitative comparisons are shown in Figure \ref{fig6} and Figure \ref{fig7}, highlighting differences in road extraction performance. Each figure includes the input image, ground truth, and outputs from LinkNet50~\cite{ref14}, U-Net~\cite{ref8}, GAMSNet~\cite{ref15}, Swin Transformer~\cite{ref23} (SwinT), SwinT+U-Net, and DB SwinT. All models were trained and tested under identical conditions, and predicted maps retain the original image resolution to capture fine-grained road structures.

On the DeepGlobe dataset, DB SwinT achieves superior performance across all metrics, demonstrating strong adaptability to complex rural environments with dense vegetation, irregular road patterns, varying widths, scattered settlements, and uneven terrain that exacerbate occlusions and structural ambiguity. It attains a Precision of 89.55\%, Recall of 87.38\%, F1-score of 88.21\%, and IoU of 79.35\%, corresponding to improvements of 10.26\% to 6.63\% in F1-score and 14.93\% to 7.04\% in IoU compared with conventional models such as LinkNet50, U-Net, GAMSNet, and SwinT. Competing methods often produce discontinuous roads, merge narrow segments, blur boundaries, or fail to recover occluded sections, particularly in areas with dense tree coverage or low-contrast roads. In contrast, DB SwinT effectively separates closely spaced roads, reconstructs occluded segments, preserves connectivity and boundary clarity, maintains accurate road topology, and achieves more consistent predictions across diverse rural scenes. These results demonstrate the robustness of DB SwinT in rural areas with low contrast, vegetation occlusion, and sparsely connected roads, while significantly improving both detection precision and structural continuity, thereby enabling more reliable large-scale road network extraction.

On the Massachusetts dataset, which features dense buildings, narrow streets, irregular intersections, frequent shadows, and overlapping urban infrastructure, DB SwinT achieves strong performance with a Precision of 86.82\%, Recall of 85.75\%, F1-score of 85.28\%, and IoU of 74.84\%. Compared with the other methods, these results correspond to F1-score improvements of 5.03\%–12.23\% and IoU gains of 6.6\%–13.25\%, demonstrating a consistent advantage across both dense urban cores and suburban regions. DB SwinT effectively handles occlusions caused by buildings, trees, vehicles, and overlapping road elements, including multilane streets, curved segments, and partially blocked lanes. Other methods frequently produce fragmented or discontinuous predictions, misclassify buildings as roads, or fail to preserve road continuity in complex intersections. By leveraging complementary features from the dual branches, DB SwinT accurately reconstructs occluded and intersecting segments, preserves fine-grained connectivity, maintains overall network consistency, and effectively handles irregular urban patterns, delivering highly accurate, coherent, and topologically correct road networks in both densely built and suburban environments.

\begin{table}
\caption{Quantitative comparison of different methods on the DeepGlobe dataset. The best results are highlighted in bold.}
\label{tab1}
\centering
\renewcommand{\arraystretch}{1.5}
\setlength{\tabcolsep}{12pt}
\begin{tabular}{lcccc}
\hline
\textbf{Method} & \textbf{Precision (\%)} & \textbf{Recall (\%)} & \textbf{F1-score (\%)} & \textbf{IoU (\%)} \\
\hline
LinkNet50   & 78.51 & 76.45 & 77.95 & 64.42 \\
U-Net       & 79.28 & 78.47 & 78.05 & 67.36 \\
GAMSNet     & 84.32 & 82.19 & 83.67 & 72.31 \\
SwinT       & 83.11 & 79.08 & 81.58 & 69.28 \\
SwinT+U-Net & 87.28 & 84.09 & 85.97 & 75.37 \\
\textbf{Ours} & \textbf{89.55} & \textbf{87.38} & \textbf{88.21} & \textbf{79.35} \\
\hline
\end{tabular}
\end{table}

\begin{table}
\caption{Quantitative comparison of different methods on the Massachusetts dataset. The best results are highlighted in bold.}
\label{tab2}
\centering
\renewcommand{\arraystretch}{1.5}
\setlength{\tabcolsep}{12pt}
\begin{tabular}{lcccc}
\hline
\textbf{Method} & \textbf{Precision (\%)} & \textbf{Recall (\%)} & \textbf{F1-score (\%)} & \textbf{IoU (\%)} \\
\hline
LinkNet50   & 76.52 & 75.26 & 75.93 & 61.59 \\
U-Net       & 78.75 & 73.54 & 76.05 & 62.79 \\
GAMSNet     & 81.43 & 80.08 & 78.31 & 68.16 \\
SwinT       & 79.28 & 76.54 & 77.03 & 64.05 \\
SwinT+U-Net & 81.58 & 79.28 & 80.25 & 68.24 \\
\textbf{Ours} & \textbf{86.82} & \textbf{85.75} & \textbf{85.28} & \textbf{74.84} \\
\hline
\end{tabular}
\end{table}

To further verify the effectiveness of the proposed dual-branch architecture, ablation experiments were conducted by comparing DB SwinT with the single-branch SwinT+U-Net under identical configurations. As reported in the last two rows of Table \ref{tab1} and Table \ref{tab2}, DB SwinT consistently achieves higher Precision and IoU on both datasets, with improvements of 2.27\% to 4.61\% in Precision and 3.98\% to 6.60\% in IoU, highlighting the contribution of multi-branch feature integration in enhancing segmentation accuracy. Visual comparisons further confirm that the dual-branch design reduces false positives, preserves road connectivity, and enhances completeness under various occlusion conditions. For example, in the second row of Figure \ref{fig6}, SwinT+U-Net misclassifies a bush as a road and partially misses narrow segments, whereas DB SwinT produces accurate segmentation results, reconstructs occluded sections, and maintains continuous road boundaries. This observation demonstrates that the dual-branch architecture effectively improves discrimination between road and non-road regions in complex and heterogeneous backgrounds, suppressing false detections while maintaining structural integrity and topological consistency under challenging occlusion and varying road conditions.

In summary, DB SwinT outperforms state-of-the-art models on rural and urban datasets. It handles complex road structures, vegetation, occlusions, narrow streets, and irregular intersections. Quantitative results show improvements in IoU, F1-score, Precision, and Recall, while qualitative analysis demonstrates better preservation of connectivity, boundary clarity, and topology. Ablation studies verify that the dual-branch design enhances feature representation, reduces false positives, and strengthens road discrimination. Overall, DB SwinT delivers accurate and robust road extraction for high-resolution remote sensing imagery.

\section{Discussion}
To explore how the number of branches affects road extraction performance, we conducted a systematic comparison of single-branch, dual-branch, and triple-branch architectures. The goal was to determine whether adding more parallel encoders consistently enhances feature representation and segmentation accuracy, or if performance gains plateau beyond a certain complexity. Based on the strong results of the dual-branch design, three configurations were evaluated: a single-branch Swin Transformer U-Net (SwinT U-Net), the proposed dual-branch DB SwinT, and a triple-branch variant (TB SwinT). Experiments were carried out on the DeepGlobe and Massachusetts datasets, representing rural and urban environments with distinct structural characteristics, occlusion patterns, and scene complexities.

The experiments utilized two widely adopted road extraction benchmarks, DeepGlobe and Massachusetts, which differ in scene composition and occlusion conditions. The single-branch architecture, SwinT U-Net, employs a single encoder to extract hierarchical features and serves as the baseline. The dual-branch DB SwinT incorporates two parallel encoders to capture complementary local and global features at multiple scales. Extending this concept, the triple-branch TB SwinT adds an extra parallel encoder to further diversify feature representations. Quantitative results comparing these architectures on both datasets are presented in Table \ref{tab3}.

\begin{table}[!b]
\centering
\renewcommand{\arraystretch}{1.4}   
\caption{Road extraction performance comparison on DeepGlobe and Massachusetts datasets.}
\label{tab3}
\begin{tabular}{@{}llllcccc@{}}
\toprule
\textbf{Dataset} & \textbf{Architecture} & \textbf{Method} & \textbf{Patch size} & \textbf{Precision (\%)} & \textbf{Recall (\%)} & \textbf{F1 (\%)} & \textbf{IoU (\%)} \\
\midrule
\multirow{13}{*}{DeepGlobe} & \multirow{4}{*}{Single-branch} & \multirow{4}{*}{SwinT U-Net} & $s=1$ & 83.62 & 82.43 & 83.02 & 70.97 \\
 & & & $s=4$ & 87.28 & 84.09 & 85.97 & 75.37 \\
 & & & \textbf{$s=8$} & \textbf{87.89} & \textbf{85.23} & \textbf{86.54} & \textbf{76.27} \\
 & & & $s=12$ & 84.65 & 82.79 & 83.71 & 71.98 \\
\cline{2-8}
 & \multirow{6}{*}{Dual-branch} & \multirow{6}{*}{DB SwinT} & $s=1,4$ & 84.45 & 82.07 & 83.27 & 71.26 \\
 & & & $s=1,8$ & 87.56 & 85.65 & 84.54 & 73.35 \\
 & & & $s=1,12$ & 87.04 & 85.21 & 84.07 & 72.52 \\
 & & & \textbf{$s=4,8$} & \textbf{89.55} & \textbf{87.38} & \textbf{88.21} & \textbf{79.35} \\
 & & & $s=4,12$ & 88.69 & 86.54 & 87.54 & 78.06 \\
 & & & $s=8,12$ & 89.03 & 87.15 & 88.16 & 78.79 \\
\cline{2-8}
 & \multirow{3}{*}{Triple-branch} & \multirow{3}{*}{TB SwinT} & $s=1,4,8$ & 81.24 & 77.62 & 79.86 & 65.85 \\
 & & & $s=1,4,12$ & 80.81 & 76.95 & 78.82 & 65.75 \\
 & & & \textbf{$s=4,8,12$} & \textbf{83.58} & \textbf{79.04} & \textbf{81.12} & \textbf{68.49} \\
\midrule
\multirow{13}{*}{Massachusetts} & \multirow{4}{*}{Single-branch} & \multirow{4}{*}{SwinT U-Net} & $s=1$ & 78.79 & 75.56 & 77.15 & 62.30 \\
 & & & $s=4$ & 81.58 & 79.24 & 80.25 & 68.24 \\
 & & & \textbf{$s=8$} & \textbf{82.35} & \textbf{81.26} & \textbf{81.77} & \textbf{69.15} \\
 & & & $s=12$ & 80.54 & 77.53 & 79.03 & 65.38 \\
\cline{2-8}
 & \multirow{6}{*}{Dual-branch} & \multirow{6}{*}{DB SwinT} & $s=1,4$ & 83.45 & 79.58 & 81.38 & 68.59 \\
 & & & $s=1,8$ & 82.36 & 78.97 & 80.52 & 67.43 \\
 & & & $s=1,12$ & 81.57 & 77.55 & 79.31 & 65.87 \\
 & & & \textbf{$s=4,8$} & \textbf{86.82} & \textbf{85.75} & \textbf{85.28} & \textbf{74.84} \\
 & & & $s=4,12$ & 85.58 & 83.26 & 84.39 & 72.98 \\
 & & & $s=8,12$ & 85.79 & 83.57 & 84.51 & 73.32 \\
\cline{2-8}
 & \multirow{3}{*}{Triple-branch} & \multirow{3}{*}{TB SwinT} & $s=1,4,8$ & 77.68 & 72.45 & 74.98 & 60.04 \\
 & & & $s=1,4,12$ & 78.84 & 73.79 & 76.43 & 61.58 \\
 & & & \textbf{$s=4,8,12$} & \textbf{79.01} & \textbf{73.87} & \textbf{76.52} & \textbf{61.62} \\
\bottomrule
\end{tabular}
\end{table}

To assess the impact of feature granularity, experiments were conducted using different patch sizes: $s=1$, $s=4$, $s=8$, and $s=12$. For the single-branch SwinT U-Net, optimal results were obtained with $s=8$. Under this setting, the model achieved an F1-score of 86.54\% and an IoU of 76.27\% on the DeepGlobe dataset, and an F1-score of 81.77\% and an IoU of 69.15\% on the Massachusetts dataset. While these outcomes demonstrate that the single-branch architecture can capture global contextual information effectively, its performance still lags behind the dual-branch design. This gap arises from the limited capacity of a single encoder to simultaneously represent global context and fine-grained local details, particularly in heavily occluded scenes. These findings underscore the importance of multi-branch architectures for addressing the inherent complexity of road extraction tasks.

\begin{figure}
\centering
\includegraphics[scale=1]{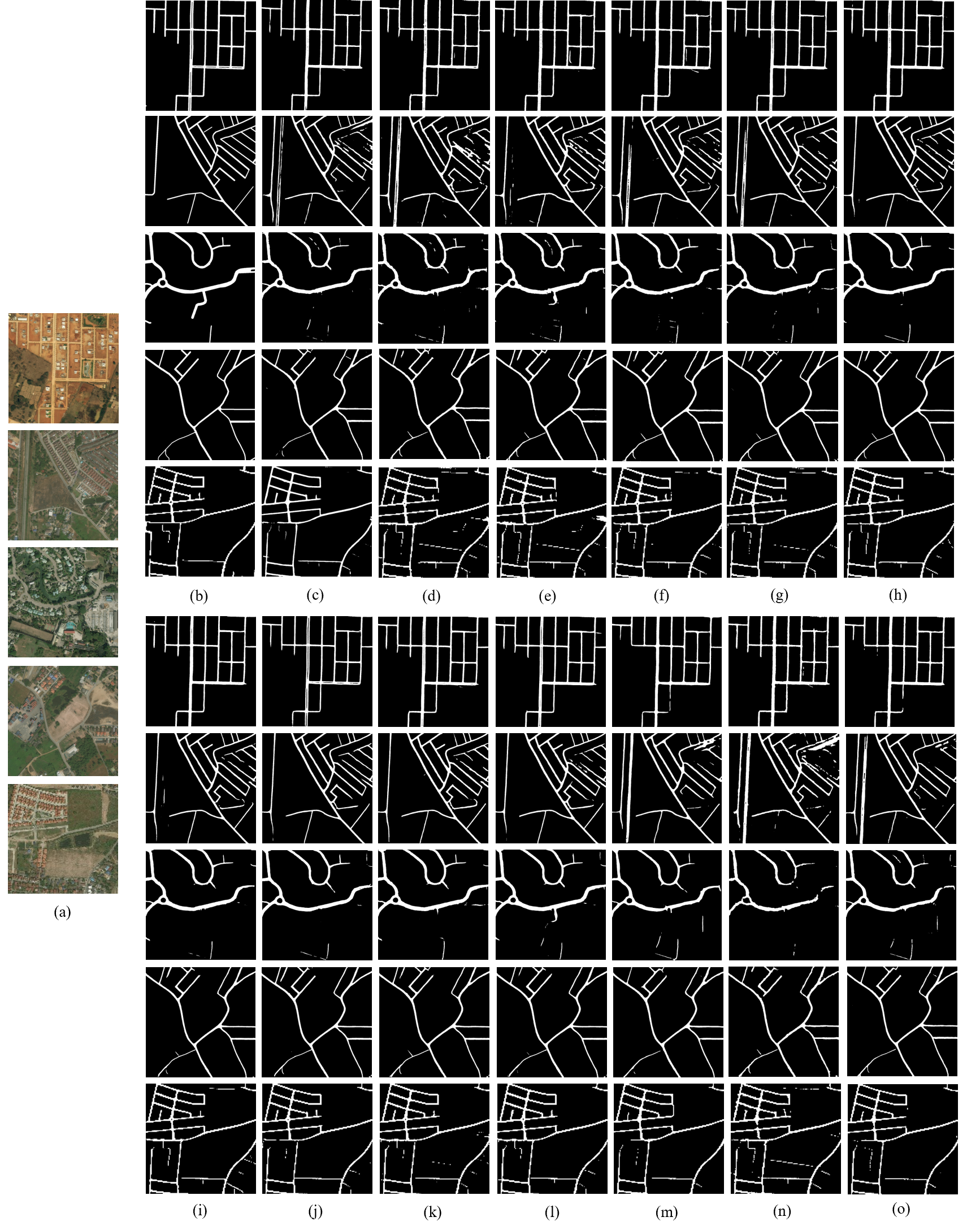}
\caption{Comparison of different architectures for road extraction on the DeepGlobe dataset. 
(a) Image. (b) Ground truth. 
(c) $1\times1$. (d) $4\times4$. (e) $8\times8$. (f) $12\times12$. 
(g) $1\times1, 4\times4$. (h) $1\times1, 8\times8$. (i) $1\times1, 12\times12$. 
(j) $4\times4, 8\times8$. (k) $4\times4, 12\times12$. (l) $8\times8, 12\times12$. 
(m) $1\times1, 4\times4, 8\times8$. (n) $1\times1, 4\times4, 12\times12$. 
(o) $4\times4, 8\times8, 12\times12$.}
\label{fig8}
\end{figure}

\begin{figure}
\centering
\includegraphics[scale=1]{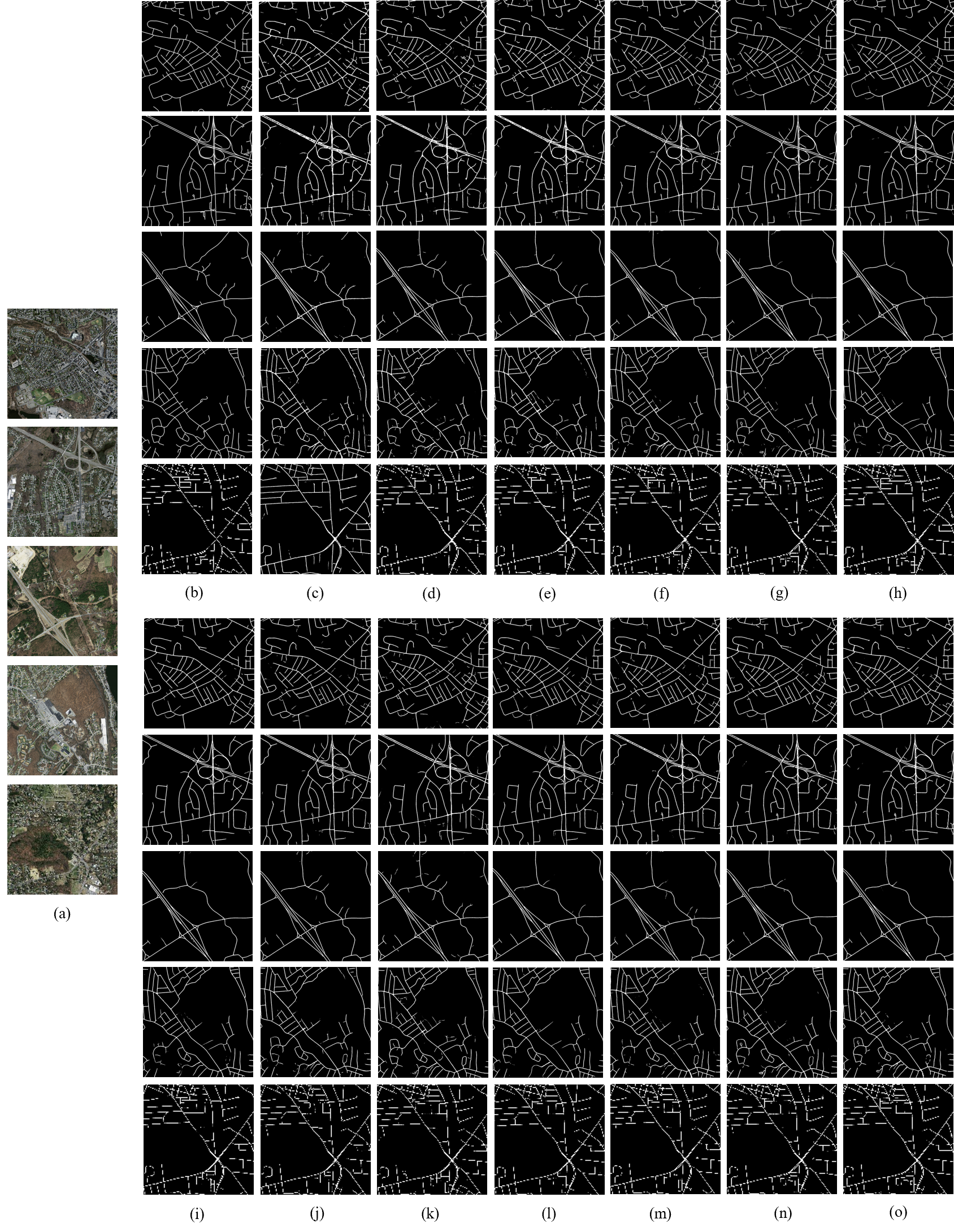}
\caption{Comparison of different architectures for road extraction on the Massachusetts dataset. 
(a) Image. (b) Ground truth. 
(c) $1\times1$. (d) $4\times4$. (e) $8\times8$. (f) $12\times12$. 
(g) $1\times1, 4\times4$. (h) $1\times1, 8\times8$. (i) $1\times1, 12\times12$. 
(j) $4\times4, 8\times8$. (k) $4\times4, 12\times12$. (l) $8\times8, 12\times12$. 
(m) $1\times1, 4\times4, 8\times8$. (n) $1\times1, 4\times4, 12\times12$. 
(o) $4\times4, 8\times8, 12\times12$.}
\label{fig9}
\end{figure}

In the second phase of experiments, the dual-branch architecture (DB SwinT) was assessed. This design employs two parallel encoders at different spatial scales, enabling the simultaneous extraction of fine-grained local features and global contextual information. Results across various patch-size combinations ($s = 1, 4, 8,$ and $12$) show that the configuration with $s = 4$ and $s = 8$ delivers the best overall performance. Specifically, DB SwinT achieves an F1-score of 88.21\% and an IoU of 79.35\% on the DeepGlobe dataset, and an F1-score of 85.28\% with an IoU of 74.84\% on the Massachusetts dataset. Compared with the single-branch model, these gains indicate that the dual-branch design more effectively balances local detail and global context representation, providing higher accuracy without significant computational overhead, and thus offers clear advantages for road extraction tasks.

The triple‑branch architecture (TB SwinT) was evaluated using patch sizes $s = 1, 4, 8,$ and $12$, with the optimal configuration achieved at $s = 4, 8,$ and $12$. Nevertheless, despite the potential advantage of an additional encoder, TB SwinT consistently performs worse than both the single‑branch and dual‑branch models in terms of F1‑score and IoU. In complex road scenes, this architecture struggles with effective feature extraction, often yielding incomplete or inaccurate road representations. Moreover, the extra computational cost of the third branch, combined with reduced robustness under occlusions, limits its practical utility. These results suggest that simply increasing the number of branches does not guarantee better performance and may introduce redundancy and optimization difficulties.

Figure \ref{fig8} and Figure \ref{fig9} show qualitative comparisons of the three architectures on the DeepGlobe and Massachusetts datasets. The DeepGlobe dataset, dominated by rural areas with substantial tree occlusions, challenges road continuity. In these cases, both the single-branch and triple-branch models often produce fragmented or blurred road predictions, especially in occluded regions. By contrast, the Massachusetts dataset primarily covers urban and suburban areas with dense buildings, where the single-branch and triple-branch architectures frequently misclassify buildings as roads, leading to inaccurate or disconnected predictions. Across both datasets, the dual-branch architecture consistently yields more complete and coherent road extractions, effectively reducing the impact of occlusions from trees, buildings, and other obstacles.
                                                                                               
Experiments on the DeepGlobe and Massachusetts datasets show that DB SwinT achieves the best overall performance across all evaluation metrics, offering superior accuracy and robustness in handling complex road boundaries and occlusions while maintaining a good balance between computational cost and representational capacity. In contrast, TB SwinT does not yield further improvement and instead exhibits a consistent performance drop. This decline can be attributed to three main factors. First, adding a third encoder increases feature redundancy, as overlapping receptive fields and similar patch scales produce highly correlated representations, limiting the effective use of complementary information. Second, fusing features from three parallel branches considerably raises alignment and integration complexity, potentially causing feature inconsistency and reducing discriminative power, especially in cluttered or heavily occluded scenes. Third, the additional structural complexity complicates optimization and increases the risk of overfitting when training data are limited. Overall, the triple-branch design shows diminishing returns, suggesting that merely increasing the number of branches does not guarantee improved road extraction performance and further supporting the effectiveness and practicality of the dual-branch architecture.

\section{Conclusion}
This study addresses the challenge of extracting roads from optical remote sensing imagery under complex backgrounds and occlusion by proposing a dual-branch Swin Transformer network, termed DB SwinT. By combining the long-range modeling capability of the Swin Transformer with the multi-scale fusion mechanism of U-Net, the architecture captures global context and fine-grained local details, while the AFF module enhances cross-branch interaction to produce continuous and precise segmentation in occluded regions. Experimental results on benchmark datasets demonstrate superior performance over alternative architectures. The dual-branch design balances global semantic understanding and local structural representation, improving representational capacity without excessive computational cost, whereas simply increasing the number of branches yields no further gains, highlighting the importance of structural efficiency and effective feature fusion. Although challenges remain in highly complex scenes, the model shows strong robustness and accuracy. Future work will improve computational efficiency for real-time and large-scale applications through lightweight Transformer variants, enhanced semantic modeling, and domain adaptation to strengthen generalization across diverse imaging conditions.

\section{Acknowledgments}
This work was supported by the National Natural Science Foundation of China Regional Innovation and Development Joint Fund Project (U22A2010). 

\bibliographystyle{unsrt}
\bibliography{references}
\end{document}